# Evaluation and Optimization of Adaptive Cruise Control in Autonomous Vehicles using the CARLA Simulator: A Study on Performance under Wet and Dry Weather Conditions

Roza Al-Hindawi
Civil Engineering Department
Al-Ahliyya Amman University
Amman, Jordan
ORCID: 0009-0002-3399-0410
202117249@ammanu.edu.jo

Taqwa I.Alhadidi
Civil Engineering Department
Al-Ahliyya Amman University
Amman, Jordan
ORCID: 0000-0001-6388-0904
T.alhadidi@ammanu.edu.jo

Mohammad Adas
Civil engineering Department
*University of Jordan*
Amman, Jordan
0000-0001-6777-3858
mhm8221368@ju.edu.jo

*Abstract*— Adaptive Cruise Control (ACC) can change the speed of the ego vehicle to maintain a safe distance from the following vehicle automatically. The primary purpose of this research is to use cutting-edge computing approaches to locate and track vehicles in real time under various conditions to achieve a safe ACC. The paper examines the extension of ACC employing depth cameras and radar sensors within Autonomous Vehicles (AVs) to respond in real-time by changing weather conditions using the Car Learning to Act (CARLA) simulation platform at noon. The ego vehicle controller's decision to accelerate or decelerate depends on the speed of the leading (ahead) vehicle and the safe distance from that vehicle. Simulation results show that a Proportional–Integral–Derivative (PID) control of autonomous vehicles using a depth camera and radar sensors reduces the speed of the leading vehicle and the ego vehicle when it rains. In addition, longer travel time was observed for both vehicles in rainy conditions than in dry conditions. Also, PID control prevents the leading vehicle from rear collisions.

*Keywords— Adaptive Cruise Control, Autonomous Vehicles, Car Learning to Act, Proportional–Integral–Derivative control.*

## I. Introduction

One of the Advanced Driver Assistance Systems (ADAS) that assumes longitudinal control of the vehicle is Adaptive Cruise Control (ACC). ACC is an intelligent cruise control technology that adjusts the vehicle's acceleration to maintain a safe space between the ego and leading vehicles. One of our day's most difficult engineering challenges is the development of autonomous vehicles. These self-driving vehicles have been projected to operate in an extremely unpredictable environment with greater dependability than humans or full autonomy. To achieve this objective, self-driving vehicles must be embedded with sophisticated algorithms and multiple sensors that allow the vehicle to recognize its surroundings in real-time regarding weather conditions. However, the vehicle must also be capable of detecting these obstacles across the scene and computing their speed.

The development and application of AVs in the transportation sector have started to be influenced by the advancements in artificial intelligence (AI) throughout time [1]. Self-driving vehicles, accompanied by enormous data produced from numerous sensors and advanced computer capabilities, have become a key element for comprehending the surrounding environment and making suitable decisions during movement. Consequently, the implementation of safe and robust Autonomous Driving (AD) holds the promise of significantly reducing road traffic accidents, congestion, and wasteful fuel consumption by transferring driving control from humans to autonomous cars. The classification system for autonomous driving, established by the Society of Automotive Engineers (SAE) and the National Highway Traffic Safety Administration (NHTSA), outlines six levels of autonomy. It begins with level 0, where the human driver is responsible for continuously monitoring all aspects of the dynamic driving task, and reaches level 5 autonomy, where the vehicle can handle all driving tasks in any situation without human intervention [2]. Given the complex and dynamic nature of urban environments, AVs must interact with various entities, such as other vehicles, pedestrians, and stationary objects [3]. To ensure safe and reliable operation, an AV must undergo rigorous safety and operational testing before deployment in real-world scenarios. Moreover, as the gap between software simulations (Such as CARLA) and real-world environments continues to narrow, trained models can be deployed on road infrastructure, providing a viable path toward realizing autonomous driving technology [4].

CARLA is utilized in this research to study the efficiency of a PID for an autonomous vehicle that consists of a vision-based perception module, a local planner, a global planner, and a traffic manager. CARLA is used to generate navigation simulations that are regulated by varying complexity. There should be control over the route's complexity, traffic flow, and the surrounding circumstances of the environment. The research results highlighted the ACC's performance characteristics in wet and dry conditions at noon on the unsignalized intersection at Town 10.

## II. Literature Review

In Asia, India, Europe, and the United States, there has been a significant concentration on AD research and development to provide creative solutions in the fields of distributed dynamic controls, computer vision, and Machine Learning (ML) as reviewed by Daily et al. [5]. However, these researchers suffered from limited ability to deal with complex problems, so the safety, reliability, and resilience of complex systems should be further investigated. Nidamanuri et al. [6] examined research gaps, reviewed ADAS functionalities, and discussed vision intelligence and computational intelligence for ADAS with learning algorithms such as supervised, unsupervised, reinforcement learning, and deep learning for real-time recommendation systems in less-disciplined road traffic. Interesting research was done by Ramakrishna et al. [7], who also suggested an ANTI-CARLA framework for automated adversarial testing, evaluation, and investigation of AV performance in the CARLA simulator. It provides a framework that makes it possible to test and plug in any AD pipeline. It consists of a straightforward interface for specifying test conditions and a domain-specific Scenario Description Language (SDL) for explaining the test conditions. The limitation of the proposed system is that it can only sample static scenes. However, the sequence in which the scenes before each unsuccessful case occurred is not now known. Gómez-Huélamo et al. [8] validated a fully self-driving vehicle architecture in the CARLA simulator using the NHTSA protocol, with an emphasis on decision-making unit analysis based on Hierarchical Interpreted Binary Petri Nets (HIBPN) and hyper-realistic simulators for real-world testing without considering the weather conditions. The architecture is tested using hard driving situations, yielding qualitative and quantitative findings for a preliminary stage before being implemented in an autonomous electric vehicle. However, the CARLA simulator is a realistic presentation device for vehicle circumstances, but there is a lack of correct communication models. Mateo et al. [9] investigated an expansion of the ms-van3t modeling framework, based on the NS-3 simulator, that combined the CARLA sensor and vehicle physics with improved communication concepts.

Gutiérrez-Moreno et al. [10] introduced a Deep Reinforcement Learning (DRL) strategy for dealing with crossings in self-driving frameworks, which combined Curriculum Learning (CL) with hostile vehicle information. The hybrid architecture was designed to solve the complexity of junctions, with an emphasis on high-level decision-making. The study examined the following situations: traffic lights, traffic signs, and uncontrolled junctions. The Proximal Policy Optimisation (PPO) algorithm predicts ego vehicle-desired behavior based on hostile vehicle behavior. There was a limitation in their research in that they obtained the state vector from real sensor information rather than directly from the actual situation. Dosovitskiy et al. [11] used CARLA to evaluate the efficiency of three AD approaches: a normal modular pipeline, an end-to-end model trained using imitation learning, and an end-to-end model trained with reinforcement learning in six weather situations. Due to computational expenses, their approach limited training to 10 million simulation steps while emphasizing failure modes. Also, Liu et al. [12] proposed an optimization-based integrated behavior planning and motion control strategy for urban AD that uses CARLA's model with Potential Functions (PFs) to characterize traffic regulations. Their model has failed collision tests in junction scenarios, traffic law violations, and roundabouts. Guo et al. [13] proposed the first end-to-end attack on ACC systems and tested the safety measures on the most advanced ACC system using Deep Neural Networks (DNNs). However, their study findings demonstrated that their approach could make a vehicle traveling with ACC accelerate unsafely, causing a rear-end collision. However, changing the vehicle's speed and communicating it through the Control Area Network (CAN) has a negative impact on ACC's Proportional-Integral-Derivative (PID) variable capabilities for crash prevention [14]. The study suggested using ML-based on real-time Intrusion Detection System (IDS) to support resilience mechanisms in mitigating cyber-attacks on vehicles and utilizing ML-based IDS to help resilience mechanisms mitigate cyber-attacks on cars.

## III. METHODOLOGY AND SETTINGS

CARLA was created as a client-server system. The server runs and displays the CARLA world. The client interface enables users to engage with the simulator by adjusting the agent's vehicle to some of the simulation features. The client consists of all the client modules used to control the actors and set up the environment conditions, which are created via the CARLA API in Python. Traffic manager is an integrated framework that uses CARLA's guidance to take control of the actors and reproduce actual urban settings. This study focused on Avs, including the development and evaluation of a system capable of handling a wide range of circumstances. The AV system equipped with ACC uses distance sensors such as a depth camera and a Radar sensor embedded in the ego vehicle and leading vehicle 1 to compute the spacing between vehicles. Also, a navigation system such as Global Positioning System (GPS) is used to locate the vehicles' locations, as illustrated in Fig.1. Using the PID control system, which is frequently employed for stability, it then executes the necessary actions, such as accelerating or decelerating, to keep a safe distance from the leading vehicles.

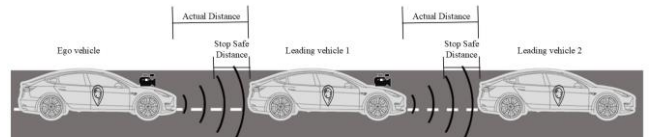

Fig. 1. Adaptive Cruise Control (ACC) model.

This model will investigate the efficiency of ACC under several environmental elements, such as the weather and time of day at Town 10 in CARLA maps. Fig.2 illustrates two case studies of severe environmental situations. The AVs start moving from the origin point to the destination point in Town 10.

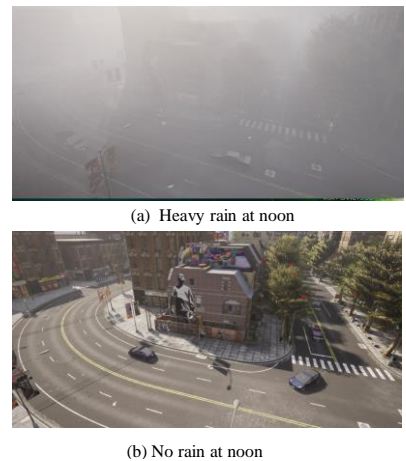

(a) Heavy rain at noon

(b) No rain at noon

Fig. 2. A street in town 10 shows two weather conditions at a curved unsignalized intersection: (a) Heavy rain at noon. (b) No rain at noon.

There are also actors in CARLA who are entities that conduct actions within the simulation and can influence other actors. They contain automobiles, pedestrians, sensors, and traffic lights. The client can change the server settings and behavior using the following commands:

- Number of vehicles: The city will have 100 non-player vehicles deployed.

- Weather ID: A list of the weather/lighting settings. The following has been studied: Noon: heavy rain, and no rain.
- Vehicle Speed: the target cruise speed is 80 km/h.
- Cameras and Radar sensors: A set of cameras with defined parameters including position, direction, and range of view. An optical depth camera is used that provides ground-truth depth segmentation. The radar sensor provides the distance, angular location, and relative speed of the obstacle.

The tested vehicle will be a Tesla Model 3, which is environmentally friendly, fast, and practical. However, there are some issues and substantial restrictions in the ADAS system that have to be solved. Tesla includes various sensors that could serve as inputs for the training model. These tough conditions need the employment of depth cameras and Radar sensors to enable safe navigation and avoid rear crashes. Also, a collision detector was used to prevent collisions between the ego vehicle and the leading vehicle, and stationary obstacle assistance was used to detect possible obstacles in front of the vehicle. A waypoint system was used to aid a vehicle reach a desired point by separating the route into multiple locations. The client provides two commands that operate the agent vehicle: (a) Throttle: pressing the accelerator pedal, represented by an integer range from zero to one. (b) Brake: pushing the brake pedal, represented by a real integer ranging from zero to one. The radar sensor provides the distance, angular location, and relative speed of the obstacle. In radar, a transmitter emits electromagnetic waves that are reflected onto the surfaces of the closest objects. The distance between the item and the sensor is estimated using the waves reflected by Equation 1.

$$D = c \times \frac{T}{2} \quad (1)$$

The equation is as follows: D represents the distance to the obstacle, c represents the speed of electromagnetic waves, and T represents the time between wave emission and reflection. The RADAR sensor outputs two-dimensional data. A scene-based approach to analysis aids in understanding complex urban processes and addressing specific urban challenges. We can examine difficulties by dividing the environment into scenes. The following scenes are under consideration: Scene (1): Unsignalized Intersection in the heavy rain at noon at a curving intersection. Scene (2): Unsignalized Intersection on a clear day at a curving intersection. we focus on the particular problems of each scenario while also analyzing AD capabilities in adaptive cruise control over 200 meters.

## IV. Suggested Method of Adaptive Cruise Controller

The PID technique is used to divide the driving task into the following subsystems: perception, planning, and continuous control. The perception layer estimates lanes, road limits, moving objects, and other risks. Furthermore, a classification algorithm is employed to determine the proximity of intersections. The local planner employs a rule-based state machine that executes simple, predefined policies tailored to urban areas. A PID controller manages the throttle and brake, providing continuous control. The definition provided by the SAE states that ACC is a Level-2 driving automation technology that adjusts the vehicle's speed based on the leading vehicle to keep a safe distance. When the distance between the vehicle and the object is less than the Stopping Safe Distance (SSD) or a collision occurs in real time, the ACC system engages emergency brakes. The safe distance is calculated using the spacing between the ego vehicle and the leading vehicle, and the ego vehicle's speed. The main objective of the ACC is to maintain a safe distance from the leading vehicle to keep the gap between the present distance $D_P$ and the safe distance D calculated using Equation 2, greater than zero.

$$D = D_p - SSD \quad (2)$$

The SSD specifies how far the vehicle drives before coming to a full stop to prevent collision with the leading vehicle, as calculated using Equation 3. where SSD is the stopping distance in meters, t is the perception-reaction time in seconds (2.5s for most drivers), v is the car's speed in kilometers per hour, G is the road's slope, and f is the coefficient of friction between the tires and the road. When the SSD is less than the present distance from the leading vehicle, the ego vehicle comes to a complete stop.

$$SSD = 0.278 \times t \times v + \frac{v^2}{254 \times f + g} \quad (3)$$

The ACC system employs two controllers, as shown in Fig.3, an upper-level controller and a lower-level controller using the CARLA simulator. The lower-level controller decides the throttle and brake, while the upper-level controller decides the appropriate longitudinal acceleration to achieve the ideal spacing and constant speed. PID controllers [15], Linear Quadratic Regulator control (LQR) [16], Sliding Mode Control [17], Fuzzy Logic Control [18], and MPC [19] are dynamic solutions proposed to implement the two-level ACC system. In this research, we use the PID model because it is simple, flexible, and relatively resilient to slow reaction times. Each controller receives the current position, speed, and list of waypoints and activates throttle and brake. The controller parameters were adjusted in Town 10.

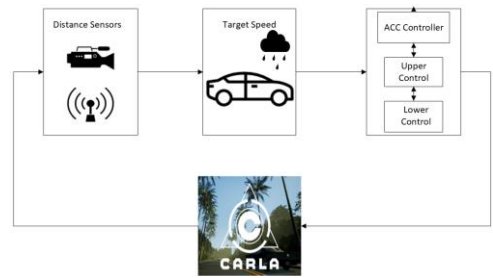

Fig. 3. Architecture design CARLA simulation with ACC.

The PID controller modifies the acceleration and deceleration to reduce the error estimated using Equation 4, which measures the difference between the vehicle's target speed (St) and its current speed (Sc) as detected by the speed sensor.

$$e(t) = S_t(t) - S_c(t) \quad (4)$$

Equation 5 employs three constants to compute the control signal u(t) based on the speed error. $K_P$ - the action's proportionate gain to the error, $K_i$ - integral gain to minimize steady-state errors by low-frequency compensation by an integrator, and $K_d$ - derivative gain to improve the transient response through high-frequency compensation by a differential [20].

$$u(t) = K_P e(t) + K_i \int_0^t e(t)\, dt + K_d \frac{d e(t)}{d_t} \quad (5)$$

Increasing the $K_P$ value allows the vehicle to reach its target speed faster, but it can exceed it. The $K_d$ term influences the reduction of the overshoot. The $K_i$ value influences the capacity to restrict steady errors and prevent fluctuations. Optimizing the gains of $K_P$, $K_d$, and $K_i$ results in a desirable overall response. On challenging driving tasks, such as curve unsignalized intersections in urbanized town 10, we employ the PID technique with two weather conditions. We employ the same agent and do not fine-tune separately for each circumstance. The tasks are set up as goal-directed navigation, with an agent starting at a specific spot in town to a specific destination. In this case, the agent isn't allowed to ignore speed limits or traffic lights. The ANOVA was used to determine whether there was a statistically significant difference in speed between the three vehicles under each weather condition.

## V. RESULTS AND DISCUSSIONS

Driving in rainy weather is frequently affected, making longitudinal control of the vehicle critical. In this paper, a longitudinal PID controller is used to implement ACC. Fig.4 illustrates the simulation environment configuration, including heavy rain and the absence of precipitation simulation situations. The three-vehicle trajectories are shown in a Three-Dimensional (3-D) graphic at noon during the time. The X-axis represents time in seconds, and the Y-axis and Z-axis represent the X and Y coordinates of the vehicles, respectively in two scenarios. The 3-D visualization provides a full perspective of how the vehicles interact with each other over the selected period. Leading vehicle 2 began to move while leading vehicle 1 stopped at the intersection to reveal the road. Also, leading vehicle 1 began to move while the ego vehicle stopped at the intersection to reveal the road, increasing the spacing to the leading vehicle 1. In the initial few seconds of the simulation, leading vehicle 2 takes a sharp right turn curve, while leading vehicle 1 follows that curve with difficulty. Also, the ego vehicle follows the leading vehicle 1 with greater difficulty in the heavy rain scenario. The plot shows that the travel time was as follows: Ego vehicle no rain: 23 seconds, leading vehicle 1 no rain: 22.5 seconds, leading vehicle 2 no rain: 11 seconds, ego vehicle heavy rain: 39 seconds, leading vehicle 1 heavy rain: 30 seconds, and leading vehicle 2 heavy rain: 24 seconds.

As a result, the travel time in ego vehicles to arrive at their destination under heavy rain increased by 69.57% when compared to no rain at noon. Furthermore, the travel time in leading vehicle 1 increased by 33.33% to arrive at the destination in heavy rain compared to no rain at noon. However, the travel time doubled leading vehicle 2 to arrive at the destination in heavy rain compared to no rain at noon. This demonstrates that under heavy rain, the vehicles take a longer time to complete their trajectories due to depth image processing.

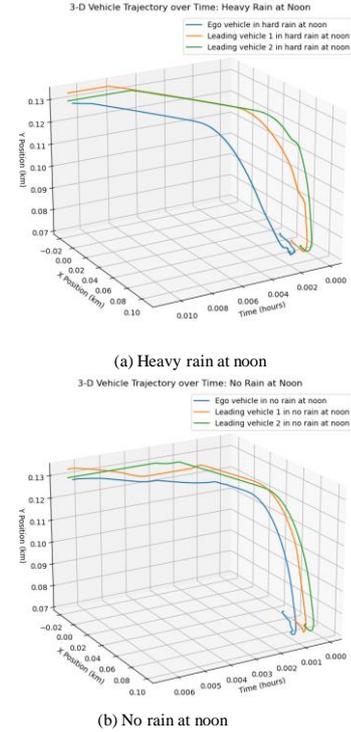

(a) Heavy rain at noon

(b) No rain at noon

Fig. 4. 3-D vehicle trajectory over time at a curved unsignalized intersection: (a) Heavy rain at noon. (b) No rain at noon.

This research focused on the speed distribution of three vehicles: the ego vehicle, leading vehicle 1, and leading vehicle 2 in heavy rain and no rain at noon. Speeds were determined using distance-recorded sensors for each vehicle. The violin-shaped plot in Fig.5 depicts the distribution of speeds for the three vehicles under the given weather circumstances.

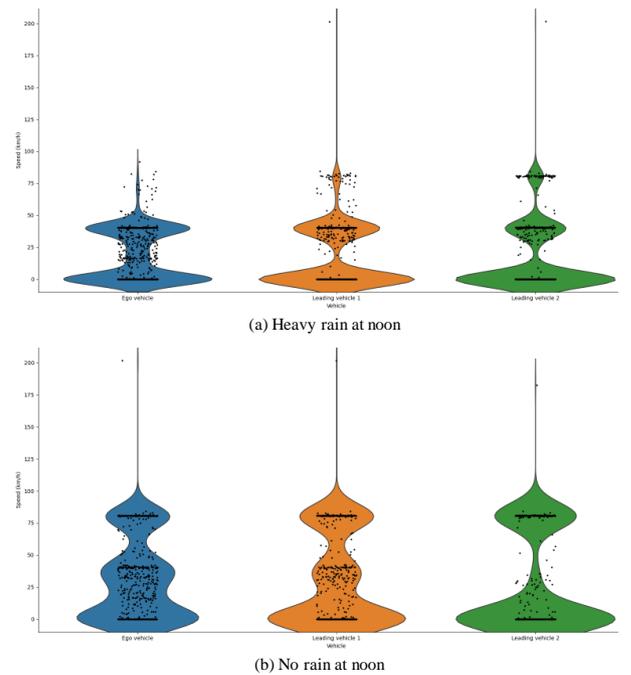

(a) Heavy rain at noon

(b) No rain at noon

Fig. 5. Speed distribution at a curved unsignalized intersection: (a) Heavy rain at noon. (b) No rain at noon.

Each violin represents the probability density of speeds, which illustrates the data set's central tendency and variability. The violin plot illustrates diverse patterns in speed distribution for the various vehicles:

- An ego vehicle in heavy rain at noon has a median speed of 14.21 km/h and an IQR of 40.31 km/h. There are potential outliers with speeds that exceed the upper whisker range, reaching 91.83 km/h. In the absence of rain at noon, the Ego vehicle has a median speed of 26.24 km/h and an IQR of 47.69 km/h. There is one possible outlier, with a speed of 201.63 km/h.

- Leading vehicle 1 in heavy rain at noon has a median speed of 0.0 km/h, indicating a distribution centered at slower speeds. The IQR is 40.32 km/h, and a large number of probable outliers are recorded, with speeds exceeding 201.15 km/h. Leading Vehicle 1 at noon in the absence of rain has a median speed of 15.46 km/h, with a wide IQR of 77.65 km/h. A probable outlier is identified at 201.62 km/h.

- Similar to leading vehicle 1, leading vehicle 2 has a median speed of 0.0 km/h in two circumstances at noon, but its IQR is 40.30 km/h in heavy rain at noon. There are numerous potential outliers, with speeds as high as 201.59 km/h. leading vehicle 2 in the absence of precipitation at noon has an IQR of 80.60 km/h, with a potential outlier of 182.56 km/h.

So, it has been observed that the decline in speed of ego vehicles between heavy rain and no rain at noon circumstances is approximately 15.46%. The drop percentage for leading vehicle 1 between heavy rain and no rain at noon is around 48.11%, at the same time for leading vehicle 2, the drop is approximately 50%. This suggests that deep image processing causes vehicles to travel slower in heavy weather. These data imply considerable variation in speed distributions, as well as the occurrence of outliers that could indicate unusual driving behavior or measurement errors. The data show differences in speed distribution among the three vehicles during heavy rain and the absence of rain at noon. The observed patterns provide useful insights into driving behavior under these particular weather circumstances.

The hypothesis was tested in both groups on the effect of different weather conditions (no rain vs. heavy rain) on the speeds of different vehicles (ego vehicle, leading vehicle 1, leading vehicle 2) in the noon. The descriptive statistics provide an overview of each vehicle's speed distribution in both weather conditions. ANOVA test was used to determine whether there was a statistically significant difference in speed between the three vehicles under each weather condition. The null hypothesis assumes that there is no significant difference among the speeds of the three vehicles. The p-value of the F-test in the no rain condition is 0.2501, demonstrating that there is no significant difference between the three vehicles' speeds. However, under heavy rain, the p-value for the F-test is 0.0494 less than $0.05$, suggesting that there are significant differences in the three vehicles' speeds. This suggests that the heavy rain has a more significant effect on vehicle speeds.

Correspondingly, temporal analysis of vehicle spacing patterns was studied over 40 seconds in both no-rain and heavy-rain circumstances as shown in Fig.6. The x-axis represents time in seconds, while the y-axis represents the distance between the ego vehicle and leading vehicle 1, between the leading vehicle 1 and leading vehicle 2, and between the ego vehicle and leading vehicle 2 (as depicted in the legend).

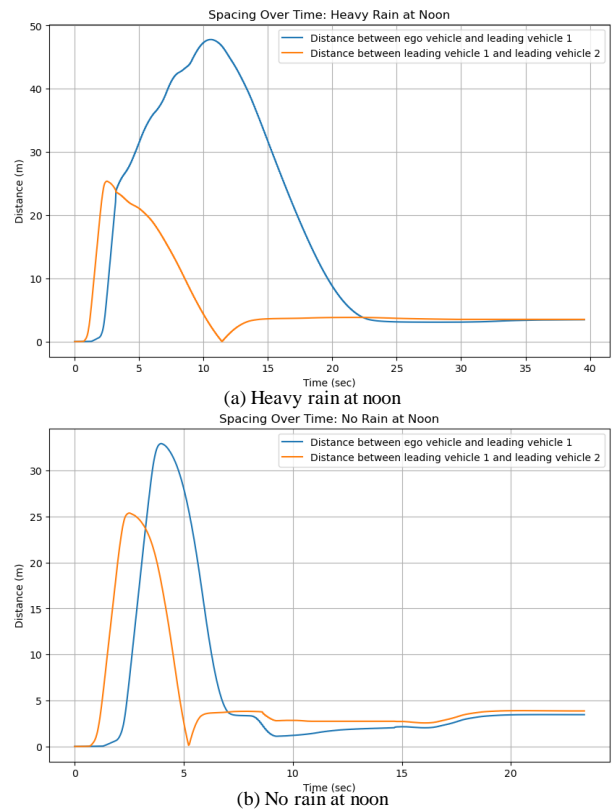

(a) Heavy rain at noon

(b) No rain at noon

Fig. 6. Spacing over time at a curved unsignalized intersection: (a) Heavy rain at noon. (b) No rain at noon.

Because of the delay at the curving unsignalized intersection, there is a distinct pattern of spacings becoming increasingly skewed to the left in both scenarios. However, the inflection point varies significantly between the two scenarios. Throughout the plot, the ego vehicle maintains a consistently shorter following distance to the leading vehicle 1. However, the leading vehicle 1 maintains a higher following distance to the leading vehicle 2 in the early seconds of the simulation in both scenarios. Compared to what happens in the absence of rain, where the three vehicles follow each other more smoothly, the following distance remains continuously shorter. In the case of no rain, the greatest effect appears in the first 7 seconds, but in the case of heavy rain, it extends to approximately 22 seconds. Then, the spacing becomes almost constant in two situations. This indicates a direct relationship between weather and spacing changes taking into consideration curved unsignalized intersections.

The hypothesis was tested in both groups to determine how varying weather conditions (no rain vs. heavy rain) affected the spacing of different vehicles (ego vehicle, leading vehicle 1, leading vehicle 2) at noon. The ANOVA test was used to determine if there was a statistically significant difference in spacing between the three vehicles in each weather condition. The null hypothesis suggests that there is no significant difference in the spacing between the three vehicles. The F-test's p-value in the no-rain scenario is 0.0022, indicating that there is a significant difference in the three vehicles' spacing distance to avoid collision. Furthermore, under heavy rain, the p-value for the F-test is 3.30e-69, which is less than the significance level of 0.05, indicating that there are significant differences in the three cars' spacing distances. In both weather conditions, there are significant distances between the vehicles. In heavy rain, the variances are more noticeable, with highly significant differences between all pairs of vehicles. The weather has a noticeable impact on spacing, and there are significant changes noted under severe rain conditions at noon.

## VI. CONCLUSION

ACC controls the vehicle's acceleration to keep a safe distance from the vehicle in front. The major goal of this research is to use edge computing technologies to locate and monitor vehicles in real-time under two weather scenarios to satisfy the ACC. The research examines the extension of ACC employing depth cameras and radar sensors within autonomous vehicles to respond in real-time changing weather conditions by modeling with the CARLA simulator with the presence of light.

Simulation findings demonstrate that AV using PID control with depth camera and radar sensors reduces the speed of the lead and ego vehicles when it heavy rain. Thus, the decrease in speed of ego vehicles at heavy precipitation at noon is roughly 15.46% compared to the no precipitation scenario. The drop percentage in the heavy rain scenario for leading vehicle 1 is around 48.11%, whereas leading vehicle 2 drops by almost 50% % compared to the no precipitation scenario. It demonstrates that in heavy weather, vehicles travel slowly due to extensive processes.

As a result, the travel time in ego vehicles to arrive at their destination during severe rain it was increased by 69.57% when compared to no rain at noon. Furthermore, the travel time in leading vehicle 1 increased by 33.33% while arriving at the destination in heavy rain versus no rain at noon. However, the travel time in leading vehicle 2 doubled to arrive at the location in rain at noon conditions. This indicates that in heavy rain, vehicles take longer to finish their trajectories due to depth computations. So, it takes longer to travel than when it is not raining to prevent the leading vehicles from rear collision. Overall, this study enhances AD research by evaluating and analyzing traffic behavior in complex urban environments.